\documentclass{esannV2}
\usepackage{graphicx}
\usepackage{subfigure}
\usepackage{amssymb,amsmath,array}
\usepackage{url}  

\begin{document}
\title{Autoregressive Convolutional Recurrent Neural Network for Univariate and Multivariate Time Series Prediction}

\author{Matteo Maggiolo and Gerasimos Spanakis
%
%
\vspace{.3cm}\\
%
Department of Data Science and Knowledge Engineering, Maastricht University\\
6200MD, Maastricht, the Netherlands%
}

\maketitle

\begin{abstract}
Time Series forecasting (univariate and multivariate) is a problem of high complexity due the different patterns that have to be detected in the input, ranging from high to low frequencies ones. In this paper we propose a new model for timeseries prediction that utilizes convolutional layers for feature extraction, a recurrent encoder and a linear autoregressive component. We motivate the model and we test and compare it against a baseline of widely used existing architectures for univariate and multivariate timeseries. The proposed model appears to outperform the baselines in almost every case of the multivariate timeseries datasets, in some cases even with 50\% improvement which shows the strengths of such a hybrid architecture in complex timeseries.
\end{abstract}

\section{Background \& Introduction\label{intro}}
Time Series (TS) modelling and forecasting has been the target of research for more than ninety years now, however the first contributions using neural networks happened in the 1980s when Recurrent Neural Networks (RNN) \cite{HopfieldNet} were introduced. New advanced variants of a base RNN were proposed in the following years, namely the Long-Short Term Memory (LSTM) network \cite{LSTMoriginal} in the 1990s and the Gated Recurrent Unit (GRU) network \cite{GRUoriginal} in 2014. These models have shown to be able to model dependencies deep in time, and therefore are very useful to model long-term, low frequency patterns in the data. In the mean time, Convolutional Neural Networks (CNN) \cite{ConvNetLecun}, initially applied to image classification were also applied to TS analysis, takes advantage of the high correlation between neighbouring time steps, since it makes the dimensional continuity assumption: features and values that are close in the input dimension (in this case, it is time) are more correlated with respect to inputs at very distant positions. Work has been done to fully exploit this type of networks in the field of TS analysis, for example by feeding in different transformation of the input \cite{MultiScaleCNN}. The continuity assumption makes this type of network very good at modelling short-term, high frequency patterns in the input signal.

Different approaches (in different fields) have tried to combine the two neural models (RNN and CNN) into a hybrid model (e.g. \cite{CNNRNNforecast}). In this direction we also contribute with a newly proposed model but also by comparing it with different baselines, namely the popular ARIMA (Auto-Regressive Integrated Moving Averages) models 
, Support Vector Machines (SVM) 
, baseline LSTM and GRU, as well as models from recent literature (and are presented in the experiments).

\section{Proposed Model}
\begin{figure*}[ht]
  \centering%
\scalebox{0.67}{
  \resizebox{\linewidth}{!}{
    \includegraphics{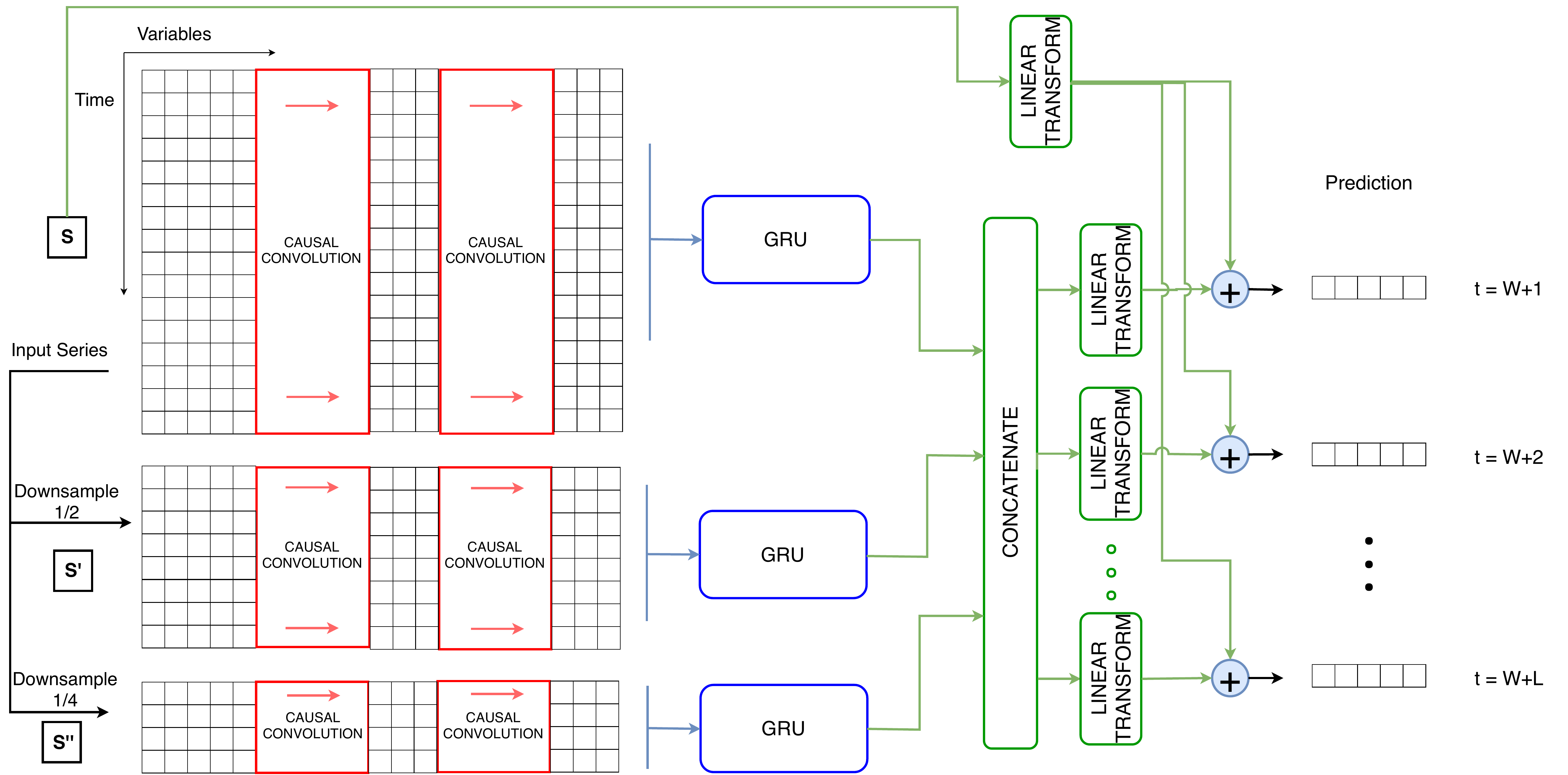}}}
    \caption{A representation of the proposed model architecture.}
\label{fig:model}
\end{figure*}

Our proposed model is inspired by the observations about RNNs and CNNs made in the Introduction and is suitable for both the univariate and multivariate TS forecasting problems. The model is composed of three parts and an overview can be seen in Figure \ref{fig:model}: (a) a multi-scale, convolutional part to extract features from the input TS, (b) a recurrent part with three GRU units to encode the sequence, followed by a linear transformation to obtain the output and (c) an autoregressive part. These parts are analyzed in detail below.

\vspace{-0.2cm}

First, the input TS is downsampled twice by a factor of 1/2 and 1/4 (averaging the values), generating a total of three input series of different lengths. This procedure will enable the convolutional layer to model a larger frequency band of the signal. In formulae, if the input TS is $S \in {R}^{T\times v}$, we have the three series:
\begin{center}
$S= (\textbf s_1,\textbf s_2,\textbf s_3,...,\textbf s_T)$\\
$
     S'=(\textbf s'_1, ..., \textbf s'_{T/2}) =\Big(\frac{\textbf s_1 + \textbf s_2}{2},\frac{\textbf s_{3} + \textbf s_{4}}{2},...\Big)$
     \\
$    S''=(\textbf s''_1, ..., \textbf s''_{T/4}) =\Big (\frac{\textbf s_1 + \textbf s_2 + \textbf s_3 + \textbf s_4}{4}, ...\Big)
$
\\
\end{center}
where we assumed that our window size $T=0 \bmod 4$. Then we apply two layers of causal convolution with the Rectifier Linear Unit (ReLU) as activation function. For filter $k=1,...,N_f$ at the first layer and for filter filter $j=1,...,N_f$ at the second layer  we will have:
$    q_k = RELU(W^{(1)}_k * S + b^{(1)}_k)
$
and
$    g_j = RELU(W^{(2)}_j * Q + b^{(2)}_j)
$.
Similarly we can define the $q$ and $g$ for the other two series (namely $q'_k$, $q''_k$, $g'_j$, $g''_j$).
where $W$ and $b$ denote the filter weights and biases, which are different for each layer and for each TS resolution, and $*$ represents a causal convolution, as described in \cite{WaveNet}. The outputs of the multi-scale convolution are the sequences $G$, $G'$ and $G''$ which have different lengths and different receptive fields over the input sequence, so as to capture multiple frequencies of patterns. We decided to use downsampling to preprocess the TS, instead of using a convolution with larger receptive field and stride, to reduce the number of parameters of the CNN layers. Furthermore, this constrains the convolutional kernels in the downsampled streams of the architecture to only model the low frequency patterns in the data. 

Next, we encode the matrices utilizing three GRU unit \cite{GRUoriginal}. We compute the GRU's hidden states as: $h_i=GRU(h_{i-1}, g_i)$ and similarly for $h'_i$ and $h''_i$. Finally, the output prediction of the nonlinear part at time $t$ is computed with a linear transformation of the concatenation of last recurrent hidden states, that is $o_t = W^{rec}_t[h_T, h'_{T/2}, h''_{T/4}] + b^{rec}_t$, where $W^{rec}_t$ and $b^{rec}_t$ are the weights/biases to learn. Note that these are different for every output time step $t$.

Finally, since the network components are nonlinear, as denoted in \cite{CNNRNNforecast}, the outputs are not sensitive to changes in the scale of the inputs, and this might give rise to problems in the case of non-stationary scale changes in the input sequences. Therefore, we subdivide the model prediction in a non-linear part (as explained in the two previous sections), plus a linear regression on the inputs. The linear prediction is given by $l^j = \mathbf{s}^j\ W^{lin} + b^{lin}$, where $W^{lin} \in R^{T\times L}$, $b^{lin} \in {R}^{L}$, $L$ is the length of the model's prediction, $\mathbf{s}^j$ is a column vector representing the input series for the variable $j$. The weights and biases for the linear shortcut are shared among the different variables. Furthermore, we reduce the regression input window to 5 previous steps in all cases.

The column vector $l^j$ is therefore the linear prediction for the variable $j$. The final prediction for the whole model is then the sum of the linear and non-linear predictions, that is $\hat{s}^j_{T+t} = o_t^j + l_t^j$. Afterwards the overall model is trained by minimizing the mean square loss function, i.e. $L= \frac{1}{N}\sum_i (Y_i - \hat{Y}_i)^2$. 


\section{Experiments}
We are testing performance for both univariate and multivariate TS forecasting, thus we are using 4 different datasets, two univariate (Daily values for Melbourne's minimum  temperature\footnote{https://datamarket.com/data/set/2324/daily-minimum-temperatures-in-melbourne-australia-1981-1990} and The Zurich Sunspot dataset\footnote{https://datamarket.com/data/set/22t4/monthly-sunspot-number-zurich-1749-1983}) and two multi-variate (Energy production for 10 different photovoltaic power plants in California \footnote{https://www.nrel.gov/grid/solar-power-data.html} and SML2010 dataset\footnote{https://archive.ics.uci.edu/ml/datasets/SML2010}, containing internal and external measurements in a domotic house).

For all datasets, we apply per-variable normalization with $\mu=0$ and $\sigma=1$. For all datasets we also apply a gaussian filter of size 5 and std 2 (for each variable), in order to smooth out some of the noise in the data. The model comparison is performed using k-fold cross-validation with $k=5$. As per metrics, we compare the model prediction's accuracy on the test set using three different metrics, the Mean Squared Error (MSE), the Mean Absolute Error (MAE) and the Dynamic Time Warping (DTW) (using FastDTW \cite{FastDTW}). 

Three types of experiments were performed with the aforementioned models. First all the models were trained and compared to predict only one step into the future, then a subset of suitable models were trained to predict more than one output step out of time, and subsequently compared them against each other and themselves with also different prediction lengths. The last experiments test the need for the autoregressive shortcut, and they were performed using synthetic data that presents varying levels of trend and periodicity.

Table \ref{table:One-step multivariate} shows the results for the univariate methods. Here it is possible to see that, for the Temperature dataset, the ARIMA model is significantly better against the baseline and the proposed model. The proposed model still ranks second, but not significantly better than the Simple LSTM model. For the Sunspot dataset, on the other hand, the original LSTNet \cite{CNNRNNforecast} performs best on average but the uncertainty on the measures suggest that further experiments should be performed to better validate these rankings. The proposed structure still ranks third on average, but it is very close in performance with the first two.

\begin{table*}[h]
\centering
\scalebox{0.7}{
  \resizebox{\linewidth}{!}{
 \begin{tabular}{|c ||c| c| c| c|}
 \hline
 Name & \multicolumn{4}{|c|}{One step forecast loss (Univariate)} \\
 \hline
Dataset & \multicolumn{2}{c|}{Temperature} & \multicolumn{2}{ c |}{Sunspot}\\ [0.5ex] 
Loss & MSE($\times10^2$) & MAE($\times10$) & MSE($\times10^2$) & MAE($\times10$) \\
 \hline\hline
 Simple LSTM 
 & 1.362$\pm$0.126 & 0.9197$\pm$0.0400
 & 0.564$\pm$0.024 & 0.5425$\pm$0.1076 \\ 
 \hline
 Deep GRU 
 & 2.003$\pm$0.220 & 1.1077$\pm$0.0554
 & 2.238$\pm$0.271 & 0.9060$\pm$0.0348 \\
 \hline
 SVM 
 & 1.771$\pm$0.338 & 1.0321$\pm$0.0837 
 & 2.773$\pm$0.455 & 0.7133$\pm$0.0332\\
 \hline
 ARIMA 
 & \textbf{1.190$\pm$0.022} & \textbf{0.8659$\pm$0.0102} 
 & 0.492$\pm$0.081 & 0.5230$\pm$0.0468\\
 \hline
 LSTNet & 1.447$\pm$0.090
 & 0.9530$\pm$0.0548 & \textbf{0.477$\pm$0.087} & \textbf{0.4980$\pm$0.0505}
\\
 \hline
 Our Model 
 & 1.317$\pm$0.083
 & 0.9019$\pm$0.0290
 & 0.501$\pm$0.126 & 0.5194$\pm$0.0653
\\
 \hline
\end{tabular}}}
 \label{table:One-step univariate}
 \caption{Results on the 1-step TS prediction task on the univariate datasets}
\end{table*}

On the other hand, for the task of multivariate prediction (results can be find in Table \ref{table:Multi-step univariate}), the proposed model is able to outperform all the other models used for comparison, comprised the LSTNet, which follows up in the second place, with a 40\% higher average error (MAE) on the Energy dataset, and a 20\% higher average error on the SML2010 dataset. These results confirm our hypothesis that for a simple prediction (just one step ahead) univariate timeseries problem, traditional models (such as ARIMA) perform very well but when multiple variables are introduced then the ability of CNNs and RNNs to capture their complex relations proves necessary for improving performance. 

\begin{table*}[h]
\centering
\scalebox{0.7}{
  \resizebox{\linewidth}{!}{
 \begin{tabular}{|c ||c| c| c| c|}
 \hline
 Name & \multicolumn{4}{|c|}{One step forecast loss (Multivariate)} \\
 \hline
Dataset & \multicolumn{2}{c|}{Energy} & \multicolumn{2}{ c |}{SML 2010}\\ [0.5ex] 
 Loss & MSE($\times10^2$) & MAE($\times10$) & MSE($\times10^2$) & MAE($\times10$) \\
 \hline\hline
 Simple LSTM 
 & 0.691$\pm$0.025 & 0.5734$\pm$0.0046
 & 1.070$\pm$0.090 & 0.5676$\pm$0.0762 \\ 
 \hline
 Deep GRU 
 & 8.563$\pm$2.239 & 2.0517$\pm$0.2572
 & 2.588$\pm$0.296 & 1.0106$\pm$0.0751 \\
 \hline
 RidgeReg & 0.626$\pm$0.175 & 0.5203$\pm$0.0717
 & 0.517$\pm$0.281 & 0.3343$\pm$0.0505
 \\
 \hline
 LSTNet 
 & 0.164$\pm$0.044 & 0.2551$\pm$0.0313
 & 0.105$\pm$0.035 & 0.1269$\pm$0.0178
 \\
 \hline
 Our Model & \textbf{0.101$\pm$0.037} & \textbf{0.1824$\pm$0.0374}
 & \textbf{0.085$\pm$0.037}
 & \textbf{0.1061$\pm$0.0223}
 \\
 \hline
\end{tabular}}}
 \label{table:One-step multivariate}
 \caption{Results on the one step TS prediction task on the multivariate datasets}
\end{table*}

\begin{figure*}[h]
  \centering%
  \scalebox{0.8}{
    \includegraphics[width=.5\linewidth]{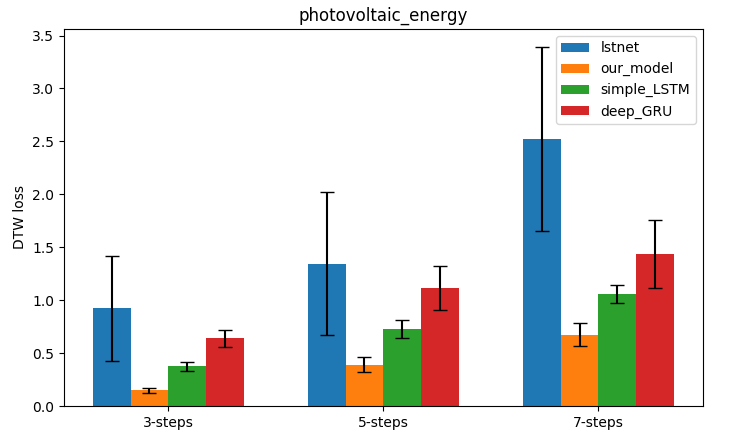}}
    \scalebox{0.8}{
    \includegraphics[width=.5\linewidth]{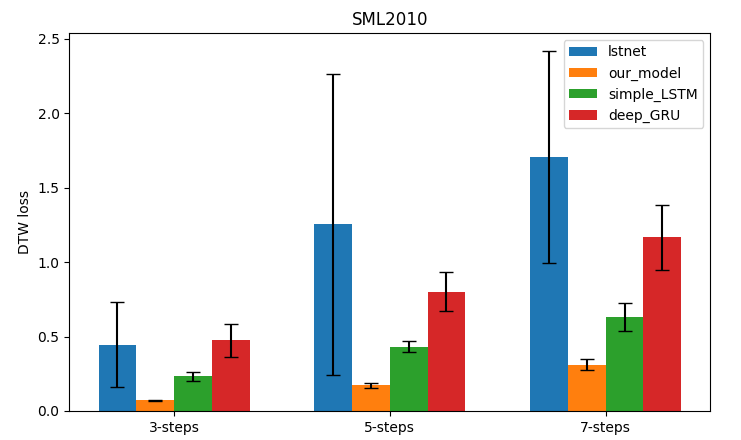}}
    \caption{Results of DTW on the multivariate datasets}
    \label{fig:multivDTW}
\end{figure*}

Finally, for the task of multiple steps prediction, results are summarized in
Table \ref{table:Multi-step univariate} (for univariate datasets) and Figure \ref{fig:multivDTW} (for multivariate). The loss displayed for both cases is Dynamic Time Warping on the output time steps. Looking at the results of univariate datasets, it is possible to see that in all cases but one the simple LSTM model is clearly better than others but also the loss values are very close and not significantly different.

The story is different for the multivariate datasets, where we are able to better distinguish the models' performances, thus we opted for a figure to better highlight the differences. In both datasets the results are similar: the suggested model significantly outperforms the baseline, and achieves a very low overall loss on the predictions. Its loss is significantly smaller than LSTM which gets the second place (more than 50\% improvement) for all multiple time steps in the SML2010 dataset, and for the three steps prediction in the Energy dataset. Furthermore, as the number of predicted time steps increases, LSTNet becomes more unstable over some of the training folds, and its loss rises, possibly because of problems with our implementation. Its loss, however, when it managed to converge to the minimum on other training folds, was still higher than that of the proposed architecture. Further experiments are required to better pinpoint the comparison between our model and the LSTNet, and to clearly assess the better performance of the former.


\begin{table*}[ht]
\centering
\scalebox{0.8}{
  \resizebox{\linewidth}{!}{
 \begin{tabular}{|c||c| c| c| c| c|c|}
 \hline
 Name & \multicolumn{6}{|c|}{Multi step DTW forecast loss (Univariate)} \\
 \hline
Dataset & \multicolumn{3}{c|}{Temperature} & \multicolumn{3}{c|}{Sunspot}\\ [0.5ex] 
 \# Steps & 3-steps & 5-steps & 7-steps& 3-steps & 5-steps & 7-steps\\
 \hline\hline
 Simple LSTM 
 & \textbf{0.592$\pm$0.033} & \textbf{1.475$\pm$0.143} & 2.679$\pm$0.303
 & \textbf{0.317$\pm$0.059} & \textbf{0.720$\pm$0.111} & \textbf{1.187$\pm$0.217} \\ 
 \hline
 Deep GRU 
 & 0.651$\pm$0.063 & 1.847$\pm$0.115 & 2.710$\pm$0.156
 & 0.364$\pm$0.147 & 0.874$\pm$0.319 & 1.421$\pm$0.448 \\
 
 \hline
 LSTNet 
 & 0.811$\pm$0.008 & 1.778$\pm$0.061 & 2.870$\pm$0.115
 & 0.377$\pm$0.080 & 0.832$\pm$0.124 & 1.365$\pm$0.224
 \\
 \hline
 Our Model 
 & 0.679$\pm$0.038 & 1.672$\pm$0.133 & \textbf{2.598$\pm$0.118}
 & 0.359$\pm$0.095 & 0.859$\pm$ 0.256 & 1.331$\pm$0.362
 \\
 \hline
\end{tabular}}}
 \label{table:Multi-step univariate}
 \caption{Results on the multiple-steps prediction task on the univariate datasets}
\end{table*}
\vspace{-0.5cm}

\begin{figure*}[ht]
  \centering%
  \scalebox{0.8}{
    \includegraphics[width=.5\linewidth]{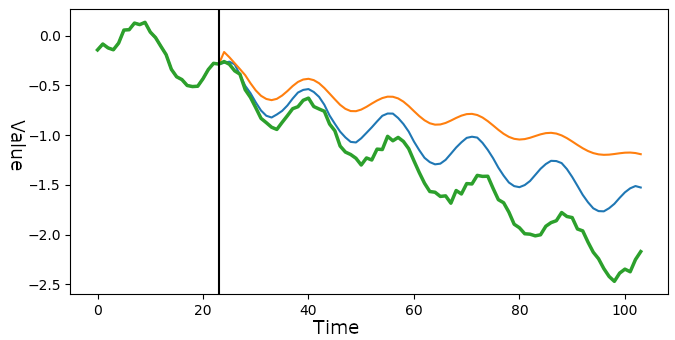}}
    \scalebox{0.8}{
    \includegraphics[width=.5\linewidth]{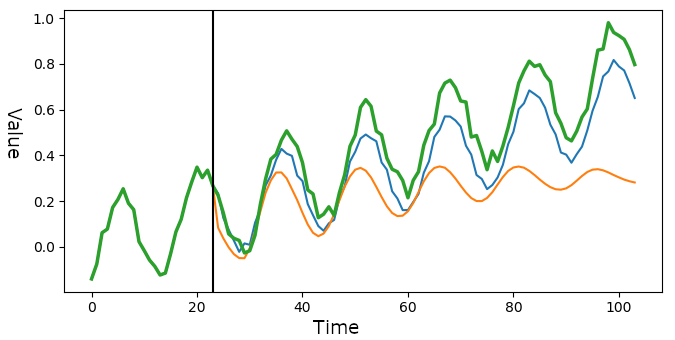}}
\caption{Comparison between model with linear regression shortcut (blue), without (orange) and the original data (green). Both models are given all  timesteps before the vertical black line and predict future time steps using a sliding window approach.}
\label{fig:ablation_study}
\end{figure*}

Finally, we perform some tests to verify the effect of the linear autoregressive shortcut. Data was generated as a list of 80 time series, each having a length of 120 steps. 
In Figure \ref{fig:ablation_study} we see the prediction error (y axis shows the value-target) using a sliding window approach. The architecture with the autoregressive component (blue) is able to better capture the variable linear trend and periodicity of the data. Without the autoregressive component (orange), the error in the prediction of the oscillations accumulates very quickly over time, while in the presence of the component the information is almost perfectly retained even after many prediction steps. With regard to the predicted trend, the model without the shortcut is noticeably worse at keeping track of the trend's slope. 
\vspace{-0.2cm}

\section{Conclusion and Future Work}
In this paper we presented a neural architecture based on three components: a feature extractor, in the form of two down-sampling levels and two convolutional layers applied to each of the inputs, a recurrent encoder, that encodes the input series and the extracted features as a fixed dimensional vector, from which the prediction will be inferred linearly and a linear connection between the inputs and the outputs that shortcuts the problem of scale insensitivity.

Detailed experiments on the model showed that a combination of a CNN and a RNN is able to outperform most of the baseline models in specific settings. More specifically, we demonstrate the effectiveness of a feature extractor over a simple recurrent encoding, especially in multivariate TS. The architecture, however, was not as effective for the task of univariate forecasting, possibly because the datasets were simple enough for simpler models. 

Further testing on the model and its hyperparameters is required to fully assess its capabilities, along with a detailed ablation study to confirm the effect of each component. It could be interesting to investigate the effect of the attention mechanism on the nonlinear structures, since it proved to be effective in sequence-to-sequence problems 
Attention modeling would give the model more freedom in which parts to look at and which parts to ignore, leading to interesting visualizations. 

\begin{footnotesize}


\bibliographystyle{unsrt}
\bibliography{mybib}

\end{footnotesize}


\end{document}